\pdfoutput=1
\documentclass[11pt]{article}

\usepackage{ACL2023}

\usepackage{times}
\usepackage{latexsym}

\usepackage[T1]{fontenc}
\usepackage{CJKutf8}
\usepackage[utf8]{inputenc}

\usepackage{microtype}

\usepackage{inconsolata}

\usepackage{booktabs}
\usepackage{siunitx} \usepackage{amssymb}
\usepackage{graphicx}
\usepackage{subfigure}
\usepackage{multirow}
\usepackage{amsmath}
\usepackage[inline]{enumitem}
\usepackage{hyperref}

\title{Are Pre-trained Language Models Useful for Model Ensemble in Chinese Grammatical Error Correction?}

\author{Chenming Tang \quad
Xiuyu Wu \quad
Yunfang Wu\thanks{\ \ \ Corresponding author.} \\
  National Key Laboratory for Multimedia Information Processing, Peking University \\
  MOE Key Laboratory of Computational Linguistics, Peking University\\
  School of Computer Science, Peking University \\
  \texttt{tangchenming@stu.pku.edu.cn} \\
  \texttt{\{xiuyu\_wu, wuyf\}@pku.edu.cn}
}

\begin{document}
\maketitle
\begin{abstract}
Model ensemble has been in widespread use for Grammatical Error Correction (GEC), boosting model performance. We hypothesize that model ensemble based on the perplexity (PPL) computed by pre-trained language models (PLMs) should benefit the GEC system. To this end, we explore several ensemble strategies based on strong PLMs with four sophisticated single models. However, the performance does not improve but even gets worse after the PLM-based ensemble. This surprising result sets us doing a detailed analysis on the data and coming up with some insights on GEC. The human references of correct sentences is far from sufficient in the test data, and the gap between a correct sentence and an idiomatic one is worth our attention. 
%On the other hand, 
Moreover, the PLM-based ensemble strategies provide an effective way to extend and improve GEC benchmark data. Our source code is available at \href{https://github.com/JamyDon/PLM-based-CGEC-Model-Ensemble}{https://github.com/JamyDon/PLM-based-CGEC-Model-Ensemble}.
\end{abstract}

\section{Introduction}
Grammatical Error Correction (GEC) is the task of automatically detecting and correcting errors in text \citep{bryant2022grammatical}. Nowadays, there are two mainstream GEC approaches. The first is treating GEC as a low-resource machine translation task \citep{yuan2016grammatical}, where sequence-to-sequence models like BART \citep{lewis2020bart} are used. This approach simply inputs the incorrect text to the encoder and gets the corrected result from the decoder. The second is treating GEC as a sequence tagging task, where the incorrect text is still taken as the input, but the output is edit tags (keep, delete, add, replace, etc.) for each token. After applying all the edits to the input text, the corrected result is then generated. The model used in this approach is also known as sequence-to-edit models and GECToR \citep{omelianchuk2020gector} is a typical one.

However, most researches on GEC focus on English while Chinese GEC (CGEC) has just started up. The Chinese language is different from English in many ways and its GEC is thus much harder. 
%For one thing, 
Instead of word inflection in many Western languages, the Chinese grammar is expressed by function words and word order, making CGEC more difficult and complex for that we can't take word form as a handle. 
%For another, 
In addition, unlike English, we have very few datasets for training and testing CGEC, which sets us exploring training-free methods like model ensemble to further improve the performance of CGEC systems.

Because of the nature of GEC that corrections can be represented as several independent edits, model ensemble has been a popular way to improve GEC systems. In CGEC, \citet{li-etal-2018-hybrid}, \citet{liang-etal-2020-bert} and \citet{zhang-etal-2022-mucgec} ensemble their models by majority voting on edits and achieve considerable improvement.
Besides, \citet{DBLP:journals/corr/XieAAJN16} adopt language models to improve neural language correction, following whom \citet{junczys-dowmunt-etal-2018-approaching} ensemble their GEC models using a language model probability. Today, transformer-based \citep{vaswani2017attention} Pre-trained Language Models (PLMs) have been in predominant
%widespread 
use in NLP. However, we find few works on model ensemble using PLMs in CGEC.

In this work, we hypothesize that choosing the best ensemble output with the help of perplexity (PPL) computed by PLMs should boost the final performance of CGEC. We experiment on ensemble of four CGEC models, including two sequence-to-sequence ones and two sequence-to-edit ones. We try four ensemble strategies: traditional voting, sentence-level ensemble, edit-level ensemble, and edit-combination ensemble, the last three exploiting the power of PLMs. 

To our surprise, the results of model ensemble with PLMs do not exceed those of traditional voting and are even worse than most of the single models. 
%To find out the causes, 
To find out why a low PPL cannot lead to a  
%dig out the relationship between low PPL and 
better GEC performance,  we carry out a detailed analysis on the ensemble results and get some insights on GEC:

1) In the test data, human references are insufficient, while PLM-based ensemble strategies produce valuable candidates, after being human checked, which may be 
considered as necessary complement to human references. 

2) When facing an erroneous sentence, a human expert corrects it with the minimal effort, while PLM-based ensemble strategies generate more natural and idiomatic text, which is of great help for oversea language learners. 

3) With the powerful ability, PLM-based models try to generate fluent sentences but sometimes ignore the original meaning of the source sentence, resulting in over-correction that should be addressed in future work.         
%Our single models and PLMs are introduced in Section \ref{section:2}. Our ensemble strategies are introduced in Section \ref{section:3}. The experimental results are shown and analyzed in Section \ref{section:4}. Case analysis is in Section \ref{sec:analysis}. The conclusion is in Section \ref{section:conclusion}.

\section{Basic Models}
\label{section:2}

\subsection{Single CGEC Models}
We implement four single models as baselines,  
%for ensemble, 
with two seq2seq models and two seq2edit ones. All the models use the Lang-8~\footnote{http://tcci.ccf.org.cn/conference/2018/taskdata.php} dataset for training. %The training datasets include Lang-8 \footnote{http://tcci.ccf.org.cn/conference/2018/taskdata.php}, HSK \footnote{https://github.com/shibing624/pycorrector},
% %\citep{Xu_Pycorrector_Text_error}, 
% THUCNews \footnote{http://thuctc.thunlp.org} 
% \citep{sun2016thuctc}, and CGED \footnote{https://github.com/blcuicall/cged\_datasets} (\citealp{rao2020overview}; \citealp{rao2018overview}), which are widely used in CGEC studies.

\paragraph{Sequence to Sequence Models.}
The two seq2seq models are both based on BART-base-Chinese
% \footnote{https://huggingface.co/fnlp/bart-base-chinese} 
\citep{shao2021cpt}, and are implemented using fairseq~\footnote{https://github.com/facebookresearch/fairseq}\citep{ott2019fairseq}. Besides Lang-8, the HSK data ~\footnote{https://github.com/shibing624/pycorrector} is also used for training. One seq2seq model adopts the "dropout-src" strategy,
% uses a strategy called "dropout-src", 
%which means 
where each token in input sentences is replaced with "[PAD]" with a probability of 10\%. The other one is pre-trained on the synthetic data constrcted on THUCNews ~\footnote{http://thuctc.thunlp.org} \citep{sun2016thuctc} before the normal training.

\paragraph{Sequence to Edit Models.}
We apply GECToR-Chinese~\footnote{https://github.com/HillZhang1999/MuCGEC} \citep{zhang-etal-2022-mucgec} as our seq2edit models, 
with the pre-trained Structbert-large-Chinese~\footnote{https://huggingface.co/bayartsogt/structbert-large} \citep{wang2019structbert} as backbone. Our two seq2edit models only differ in random seeds.

\subsection{Pre-trained Language Models}
\label{subsec:plms}
We adopt three PLMs to carry out model ensemble.
%by consulting them about PPLs: 

\paragraph{BERT-base-Chinese \footnote{https://huggingface.co/bert-base-chinese}.}
It is pre-trained on two tasks: Masked Language Model (MLM) and Next Sentence Prediction (NSP). In MLM, each token has a chance of 15\% to be replaced with a "[MASK]" (80\%), a random word (10\%), or itself (10\%). Please refer to \citet{devlin-etal-2019-bert} for details.

\paragraph{MacBERT-base-Chinese \footnote{https://huggingface.co/hfl/chinese-macbert-base}.}
It is similar to BERT, but employs whole word masking, N-gram masking and similar word replacing in MLM. Besides, Sentence-Order Prediction (SOP) is exploited instead of NSP. Please refer to \citet{cui-etal-2020-revisiting} for details.

\paragraph{GPT2-Chinese \footnote{https://github.com/Morizeyao/GPT2-Chinese}.}
It is an unofficial Chinese version of GPT-2 \citep{Radford2019LanguageMA}. It employs generative pre-training, 
by predicting the next word in a sentence with only previous words provided.

\section{Ensemble Strategy}
\label{section:3}

%\begin{figure*}[!t]
%\centering
%\includegraphics[scale=.4]{asserts/images/diagram.png}
%\caption{Diagram of our ensemble strategies using PLMs.}
%\label{fig:diagram}
%\end{figure*}

\begin{figure*}
\centering
\includegraphics[scale=.45]{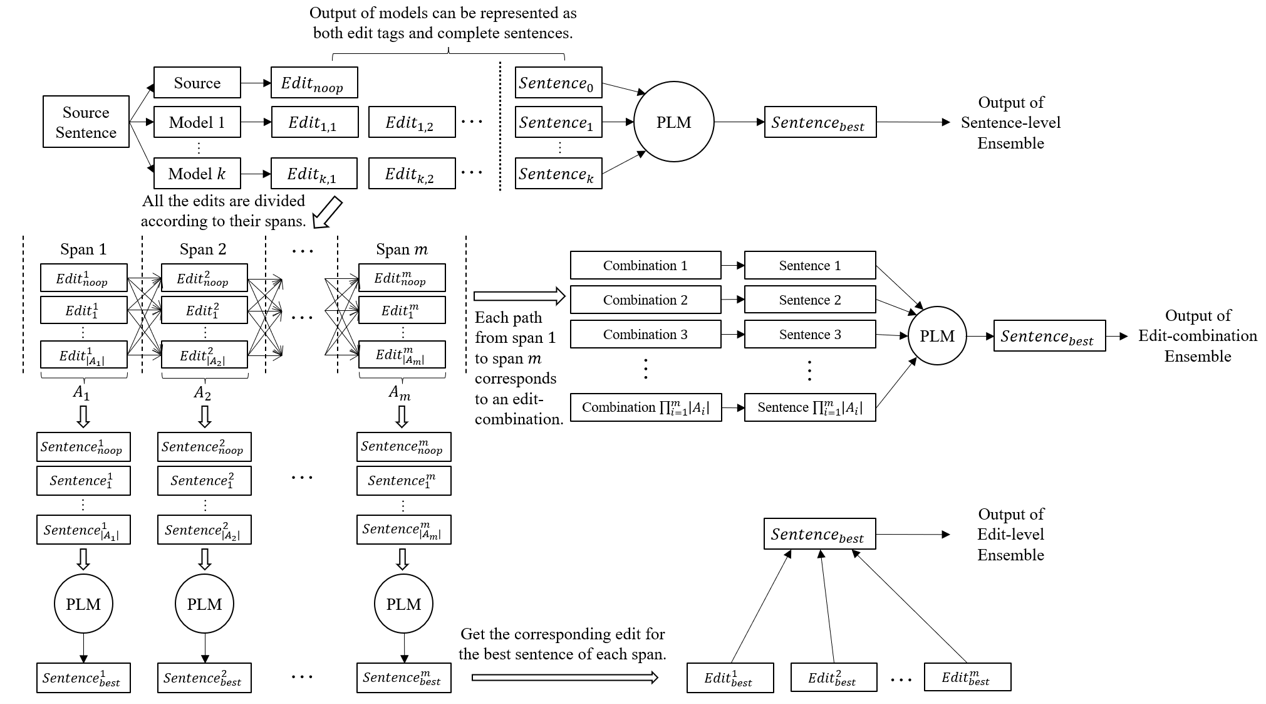}
\caption{Diagram of our PLM-based ensemble strategies.}
\label{fig:diagram}
\end{figure*}

%To investigate the effect of PLMs on CGEC model performance, 
With the source sentence and the outputs of four single models as the input, we present four ensemble strategies. 
%The basic 
%One strategy is  
%The first is 
%the traditional voting method,
%the same as \citet{zhang-etal-2022-mucgec}, 
%while the other three are designed to
%make use of PLMs’ powerful ability to decide the final output.
%For more details, please refer to Appendix \ref{sec:appendix} and see the diagrams of our ensemble strategies. 
%are shown in Figure \ref{fig:diagram}.
The diagram of our PLM-based ensamble strategies is shown in Figure \ref{fig:diagram}.

\subsection{Traditional Voting}
Different models vote for the final results.
% As \citet{zhang-etal-2022-mucgec} did, we convert all the output sentences of single models into an M2 format (\citealp{bryant-etal-2017-automatic}; \citealp{felice-etal-2016-automatic})  to get their corresponding edits, in which for each output sentence there is one line of the source sentence, one line of the output sentence, and one or more lines of edit tags. An edit tag is made up of its edit span, its error type (including missing, redundant, substitution, word-order and "noop" if the output sentence remains the same as the source one), the new tokens after the edit in the span, and some other information.
%Specifically, 
For each sentence, we consider 
%only apply 
edit operations suggested by no less than $T$ models as the correct one. 
%where $T$ is a threshold to be explored. 
In our work, we experiment on $T$ from 2 to 4.	
We implement the original code provided by \citet{zhang-etal-2022-mucgec} to carry out this voting strategy. 
%For more details, please refer to \citet{zhang-etal-2022-mucgec}.

\subsection{Sentence-level Ensemble}
%Based on 
Using different PLMs, we compute the perplexities (PPLs) of the source sentence and 
%hypothesis sentences 
the outputs
of four single models.
%(or, in other words, the highest language model probability) 
Specifically, given a sentence $S=(w_1,w_2,...,w_n)$ and the probability of the word $w_i$ computed by a PLM denoted as 
%being 
$p_i$, 
%then masked 
% generated probability of word $w_i$ in the sentence computed by bidirectional PLMs (BERT and MacBERT in this work) is:
% \begin{equation}
    % p_i=p(w_i \vert w_1w_2...w_{i-1}[MASK]w_{i+1}...w_n),
% \end{equation}
% and by generative PLMs (GPT2-Chinese) is:
% \begin{equation}
    % p_i=p(w_i \vert w_1w_2...w_{i-1}),
% \end{equation}
% PPL = \sqrt[n] { \prod_{i=1}^{n}{\frac{1}{p_i}} }.
then $PPL = (\prod_{i=1}^{n}{\frac{1}{p_i}})^{1/n}.$
The sentence with the lowest PPL is chosen to be the final output.

\subsection{Edit-level Ensemble}
\label{section:edit-level}
Given a source sentence $S$, all the edits suggested by single models constitute a candidate set $A$, and the number of edit spans is denoted as $m$. An edit span means the start-end pair of an edit's position in the sentence.
The set of all the edits (from different single models) on the $i$-th edit span (including "noop") is denoted as $A_i$. Thus, we can divide $A = \bigcup_{i=1}^{m}{A_i}$,
%\begin{equation}
%\label{eq:editset}
%    A = \bigcup_{i=1}^{m}{A_i},
%\end{equation}
where $A_i = \{ e^i_{j} \mid j = 1, 2, ..., \vert A_i \vert\}$, and $e^i_{j}$ means the $j$-th edit on the $i$-th edit span.

For each edit span 
%(e.g., the $i$-th) in $S$ (in other words, 
($A_i$ in $A$), we generate $\vert A_i \vert$ new sentences, each corresponding to a single edit in $A_i$. Then we consult PLMs about PPLs of these new sentences and accept the edit corresponding to the sentence with the lowest PPL, which we mark as $e^i_{best}$. In other words, $e^i_{best}$ is the best edit (decided by PLMs) in $A_i$, or on span $i$.

With each span's best edit, the final edit set $E_{final}$ combines these best edits, 
%and can be 
described as:
\begin{equation}
    E_{final} = \{e^i_{best} \mid i\in\{1, 2, ..., m\}\},
\end{equation}
The final hypothesis sentence is then %generated 
produced on the basis of 
$E_{final}$.

\subsection{Edit-combination Ensemble}
\label{sec:edit-combination}
One source sentence may contain more than one errors. 
%If we just rate sentences generated in Section \ref{section:edit-level}, there may be other errors in them affecting PLMs’ performance of computing PPLs. To minimize this effect, 
For each sentence, this strategy applies all edit combinations 
%(which means “noop” along with all other edits suggested by single models on each edit span are traversed and we combine edits on all edit spans) 
to the source sentence and generates many new sentences. 

To be specific, given a source sentence $S$, the edit candidates $A$ are still divided as $A = \bigcup_{i=1}^{m}{A_i}$, 
%Equation \ref{eq:editset}, 
and then we get all possible edit-combinations by:
\begin{equation}
U = \{\{e^1_{j_1}, e^2_{j_2}, ..., e^m_{j_m}\}\mid j_i \in \{1,2, ..., |A_i|\}\}.
% \begin{aligned}
    % U = \{ E_{(c_1, c_2, ..., c_m)} \mid c_i \in \{1, 2, ..., \vert A_i \vert \}, \forall i \},\\
% \end{equation}
% \begin{equation}
    % E_{(c_1, c_2, ..., c_m)}=\{e^i_{c_i} \mid i\in \{ 1, 2, ..., m\}\}.
% \end{aligned}
\end{equation}
Thus we generate ($\prod_{i=1}^{m}{\vert A_i \vert}$)
%\begin{equation}
%\label{eq:complexity}
%    \prod_{i=1}^{m}{\vert A_i \vert}
%\end{equation}
new sentences, each corresponding to an edit-combination in $U$. 
% It's worth noting that for all $i$ from $1$ to $m$, $\vert A_i \vert$ is no less than $2$ because there should be at least a single edit and a "noop" in $A_i$.
The sentence with the lowest PPL 
%in all the $\prod_{i=1}^{m}{\vert A_i \vert}$ sentences 
will be accepted as the final output.

%This strategy improves the probability of PLMs receiving perfect sentences, and thus we hypothesize that it could achieve great performance. 
Taking the %exponential 
computational complexity 
%(which can be seen in Equation \ref{eq:complexity}) 
into consideration, we only apply this strategy on sentences whose number of edit-combinations
%($\prod_{i=1}^{m}{\vert A_i \vert}$) 
is no more than 300. 
%in our work, which we call "simple sentences". 
Such simple sentences make up 95.15\% of MuCGEC-test and 98.90\% of NLPCC-test. We do nothing to the left not-so-simple sentences.

\section{Experiments}
\label{section:4}

\begin{table*}[!t]
\small
\centering
\begin{tabular}{lccccccccc}
\toprule
\multirow{2}*{\textbf{Strategy}} & \multicolumn{3}{c}{\textbf{MuCGEC-test}} & \multicolumn{3}{c}{\textbf{NLPCC-test}} & \multicolumn{3}{c}{\textbf{NLPCC-test (word-level)}}\\
\cmidrule(lr){2-4}\cmidrule(lr){5-7}\cmidrule(lr){8-10}
& \textbf{P} & \textbf{R} & $\mathbf{F_{0.5}}$ & \textbf{P} & \textbf{R} & $\mathbf{F_{0.5}}$ & \textbf{P} & \textbf{R} & $\mathbf{F_{0.5}}$\\
\toprule
{\textbf{Single Models}}\\
\midrule
seq2seq-1 & \textbf{55.00} & 28.32 & \textbf{46.28} & \textbf{43.93} & 28.21 & \textbf{39.52} & \textbf{46.17} & 29.51 & \textbf{41.48}\\
seq2seq-2 & 50.62 & 30.40 & 44.68 & 40.79 & \textbf{29.59} & 37.92 & 43.40 & 31.29 & 40.28\\
seq2edit-1 & 45.80 & 28.41 & 40.81 & 38.42 & 26.79 & 35.35 & 43.08 & 30.05 & 39.64\\
seq2edit-2 & 45.45 & \textbf{30.45} & 41.37 & 36.19 & 28.15 & 34.24 & 41.41 & \textbf{31.58} & 38.98\\
\midrule
\textit{Average of 4} & \textit{49.22} & \textit{29.40} & \textit{43.29} & \textit{39.83} & \textit{28.19} & \textit{36.76} & \textit{43.52} & \textit{30.61} & \textit{40.10}\\
\toprule
{\textbf{Traditional Voting}}\\
\midrule
% $T = 1$ & 40.25 & \underline{\textbf{37.32}} & 39.63 & 31.62 & \underline{\textbf{35.88}} & 32.39 & 38.22 & \underline{\textbf{40.73}} & 38.70\\
$T = 2$ & 52.58 & \textbf{33.61} & 47.25 & 42.71 & \textbf{32.62} & 40.22 & 45.58 & \textbf{34.66} & 42.88\\
$T = 3$ & 69.10 & 21.68 & \underline{\textbf{48.07}} & 60.81 & 21.00 & \underline{\textbf{44.09}} & 58.39 & 21.55 & \underline{\textbf{43.52}}\\
$T = 4$ & \underline{\textbf{76.13}} & 15.35 & 42.48 & \underline{\textbf{67.33}} & 14.96 & 39.61 & \underline{\textbf{64.51}} & 15.35 & 39.32\\
\toprule
{\textbf{Sentence-level}}\\
\midrule
BERT-base-Chinese & \textbf{48.56} & 24.33 & 40.50 & 37.71 & 22.80 & 33.35 & 41.38 & 24.55 & 36.39\\
MacBERT-base-Chinese & 46.83 & 33.35 & 43.33 & 37.62 & 31.30 & 36.16 & \textbf{42.24} & 34.15 & 40.33\\
GPT2-Chinese & 47.36 & \underline{\textbf{35.01}} & \textbf{44.24} & \textbf{37.75} & \textbf{33.20} & \textbf{36.74} & 41.94 & \textbf{36.13} & \textbf{40.63}\\
\toprule
{\textbf{Edit-level}}\\
\midrule
BERT-base-Chinese & 41.31 & 21.79 & 35.04 & 33.19 & 20.59 & 29.57 & 36.69 & 23.24 & 32.89\\
MacBERT-base-Chinese & 43.40 & 29.19 & 39.55 & \textbf{35.38} & 28.42 & 33.73 & \textbf{40.07} & 32.87 & 38.39\\
GPT2-Chinese & \textbf{43.93} & \textbf{33.36} & \textbf{41.31} & 35.04 & \textbf{31.60} & \textbf{34.29} & 39.44 & \textbf{36.07} & \textbf{38.71}\\
\toprule
{\textbf{Edit-combination}}\\
\midrule
BERT-base-Chinese & 42.90 & 20.18 & 35.01 & 34.25 & 21.56 & 30.64 & 37.56 & 23.94 & 33.72\\
MacBERT-base-Chinese & 45.18 & 28.73 & 40.54 & \textbf{36.35} & 30.69 & 35.05 & 40.11 & 33.62 & 38.62\\
GPT2-Chinese & \textbf{46.07} & \textbf{31.92} & \textbf{42.32} & 36.23 & \underline{\textbf{33.29}} & \textbf{35.60} & \textbf{40.50} & \underline{\textbf{36.44}} & \textbf{39.62}\\
\bottomrule
\end{tabular}
\caption{\label{tab:results}
Experimental results on MuCGEC-test and NLPCC-test.
%~\footnote{http://tcci.ccf.org.cn/conference/2018/taskdata.php}. 
%The first two columns show char-level results while the third shows word-level results on NLPCC-test. %The results of the average of single models are reported in \textit{italic}, 
The relatively best results in a group are reported in \textbf{bold}, and the best results of all are listed in \underline{\textbf{underlined bold}}.
}
\end{table*}

\subsection{Dataset and Evaluation Metrics}
\label{subsec:dataset-and-evaluation-metrics}
We carry out experiments on MuCGEC test data
%~\footnote{https://tianchi.aliyun.com/dataset/131328} 
~\citep{zhang-etal-2022-mucgec} 
and NLPCC test data \citep{zhao2018overview}.
%~\footnote{http://tcci.ccf.org.cn/conference/2018/taskdata.php}.
MuCGEC contains 7063 sentences and each have at most three references, but is not available at present. 
%So all our analysis is based on 
NLPCC contains 2000 sentences, each with one or two references, and about 1.1 references on average. We carry out analysis on NLPCC test data.

On MuCGEC, we 
submit the results of our systems to the public evaluation website~\footnote{https://tianchi.aliyun.com/dataset/131328}. 
On NLPCC, we implement the
tools provided by \citet{zhang-etal-2022-mucgec} 
%are used 
to compute the P (Precision), R (Recall), and $F_{0.5}$ of the output on char-level. Also, we report word-level results 
on NLPCC-test 
for reference 
%a fair comparison 
with previous works. 
%~\footnote{Please refer to Appendix \ref{app:nlpcc-word} for word-level results on NLPCC-test, which we don't use because of the inevitable effect of tokenizers.}. $F_{0.5}$ is our focus.

\subsection{Results}
\label{subsec:results}
Table \ref{tab:results} shows the experimental results. The traditional voting strategy achieves the best performance, with a 44.09 $F_{0.5}$ score on char level that is significantly higher than the best single model.  
%paragraph{Traditional Voting.}
With the threshold $T$ increasing, the precision rises while the recall drops.
%goes down, 
%which corresponds to common sense. 
When $T = 3$, 
%which is the number of single models minus 1, 
$F_{0.5}$ score reaches the peak,
in line with the finding of \citet{tarnavskyi2022ensembling}.

However, the PLM-based ensemble strategies get much worse performance than the simple voting strategy, and are even lower than %the best 
most of single models. In terms of precision and recall, traditional voting achieves higher precision but lower recall than single models while PLM-based strategies are on the contrary. Among three ensemble strategies, the sentence-level one performs best. 
%obtains the best result. 

%\paragraph{Sentence-level Ensemble.}
%Among the three PLMs, GPT2-Chinese achieves the highest $F_{0.5}$, 
%which is lower than half of the single models. This is surprising. We think the fact that sometimes one single model corrects one error while another single model corrects another error may lead to PLMs receiving several partly correct sentences, and thus they cannot decide the (relatively) best one, causing the decline in performance.

%\paragraph{Edit-level Ensemble.}
%The highest $F_{0.5}$ is achieved by GPT2-Chinese again, much lower than the average of single models. We think, again, the fact that sometimes PLMs receive partly correct sentences may have negative effect on their performance. In addition, 
%Edit-level Ensemble, this strategy, in a sense, is just like choosing several \textbf{locally} optimal edits and combining them into the final result, where the \textbf{global} optimum cannot be guaranteed.

%\paragraph{Edit-combination Ensemble.}
%Still, the highest $F_{0.5}$ is achieved by GPT2-Chinese and is lower than the average of single models. This sets us thinking about the relationship between PPL and $F_{0.5}$, which will be discussed in Section \ref{sec:case-analysis}.

Among different PLMs, GPT2-Chinese achieves the best results in all three ensemble strategies. %It's not difficult to explain 
This may be because BERT-based models are naturally good at mask prediction rather than computing PPLs for whole sentences. Later, we base GPT2-Chinese to make further analysis.

%When it comes to precision and recall, we can see that traditional voting achieves higher precision but lower recall than single models while PLM-based strategies are on the contrary. The former is conservative while the latter are radical when choosing corrections. However, because of the feature of $F_{0.5}$, precision-recall trade-off has to give priority to precision, which explains the superiority of traditional voting strategy in terms of CGEC.

% When it comes to Precision and Recall, we can see that after the ensemble with PLMs, the Precision goes extremely low, while Recall is not that bad. There is a situation where the PLMs have to choose among edit spans in source sentence and several generated (or, in sentence-level ensemble, outputted) sentences, all with errors on that span, since sometimes single models make errors; and PLMs cannot distinguish between these errors, not knowing which one should be the best. Theoretically, the probability that PLMs choose the source one is $\frac{1}{N}$, where $N$ is the number of candidate sentences. However, as we know, if the source one is chosen, Precision is not affected; but if one of other generated (or outputted) spans is chosen, Precision will decline. As a result, Precision will go down for nothing with a high probability in such situations, which may be responsible for the low Precision in the results.

\section{Analysis and Discussion}
%\section{Why Can't Low PPL Cause Better Performance for GEC?}
\label{sec:analysis}
We design three ensemble strategies to choose the sequence with the lowest PPL as the final output, but why does $F_{0.5}$ score drop? In our work, all single models are made up of their own PLMs, which means ensembling them exploiting another PLM is just like using PLMs to judge PLMs, so the performance may benefit little. This is in line with the work of \citet{junczys-dowmunt-etal-2018-approaching}, where pre-trained single models gain little and even have worse performance after PLM-based ensemble while other simple single models benefit a lot. Besides this, are there any other reasons?    

\subsection{Statistical Results}
\label{subsec:statistical-results}
In order to find out the cause of the poor performance of PLM-based ensemble strategies, on NLPCC test data, we randomly select 200 samples from the results of all the three strategies along with the best single model (seq2seq-1) for comparison,
and ask two graduate students to analyze the output sentences with a double-blind manner. After that, a third expert arbitrates for the inconsistency. Instructions for human annotators are shown in Appendix \ref{sec:instructions}.
%in detail 

According to human judgement, four types are summarized. 
%as shown in Table \ref{tab:statistical-results}. 
\textbf{Exact} (\textbf{E}): the output is fluent and correct, in line with the reference. 
\textbf{Good} (\textbf{G}): the output is fluent and correct but different with the reference, which indicates that the references are not sufficient enough. 
\textbf{Over-corrected} (\textbf{O}): the output is fluent but doesn't meet the original meaning of the source sentence. 
\textbf{Wrong} (\textbf{W}): the output has other problems that we don't care in this work.

The result of human annotation is reported in Table \ref{tab:statistical-results},
%is the outcome of a third-party arbitration after double-blind human annotating by two separate annotators.
and some examples of \textbf{G} and \textbf{O} are shown in Table \ref{tab:examples}.

\begin{table}[h]
\small
\centering
\vspace{-0.0em}
\begin{tabular}{lcccc}
\toprule
 & \textbf{E} & \textbf{G} & \textbf{O} & \textbf{W} \\
\toprule
seq2seq-1 (best single) & 38 & 42 & 9 & 111 \\
\midrule
Sentence-level & 36 & 53 & 23 & 88 \\
Edit-level & 32 & 45 & 20 & 103 \\
Edit-combination & 32 & 59 & 21 & 88 \\
\bottomrule
\end{tabular}
\caption{\label{tab:statistical-results}
Human annotation of generated outputs.}
%The number of different types.}
\vspace{-0.0em}
\end{table}

\begin{table}[h]
\footnotesize
% \scriptsize
%\vspace{-0.8em}
\centering
\begin{tabular}{c|l}
\hline
\multirow{4}{*}{\textbf{G}} & src: \begin{CJK}{UTF8}{gbsn}\underline{我的家}附近有很多\underline{考式}补习班。\end{CJK}                                                    \\
                    % &                      & There are many cram schools for \underline{exam form} near \underline{the home of me}.    \\
                    % \cline{2-3}
                    & out: \begin{CJK}{UTF8}{gbsn}\underline{我家}附近有很多\underline{考试}补习班。\end{CJK}                          \\
                    % &                      & There are many cram schools for \underline{examination} near \underline{my home}.         \\
                    % \cline{2-3}
                    & ref: \begin{CJK}{UTF8}{gbsn}\underline{我的家}附近有很多\underline{考试}补习班。\end{CJK}                                                    \\
                    % &                      & There are many cram schools for \underline{examination} near \underline{the home of me}.  \\
                    % \cline{2-3}
                    & \ \ \ \ \ \ \ There are many cram schools near my home.\\
\hline
\multirow{4}{*}{\textbf{G}} & src: \begin{CJK}{UTF8}{gbsn}我\underline{低幼儿童的时候}很想养狗。\end{CJK}                                                     \\
                    % &                      & I really wanted a dog \underline{at the time when I young child}.             \\
                    % \cline{2-3}
                    & out: \begin{CJK}{UTF8}{gbsn}我\underline{小时候}很想养狗。\end{CJK}                                                         \\
                    % &                      & I really wanted a dog \underline{when I was young}.                           \\
                    % \cline{2-3}
                    & ref: \begin{CJK}{UTF8}{gbsn}我\underline{小的时候}很想养狗。\end{CJK}                                                        \\
                    % &                      & I really wanted a dog \underline{at the time when I was young}.               \\
                    % \cline{2-3}
                    & \ \ \ \ \ \ \ I really wanted a dog when I was young.\\
\hline
\multirow{4}{*}{\textbf{G}} & src: \begin{CJK}{UTF8}{gbsn}可它的表情是从来没\underline{看}过的。\end{CJK}                                                     \\
                    % &                      & But it has a look have never \underline{looked at} before.                    \\
                    % \cline{2-3}
                    & out: \begin{CJK}{UTF8}{gbsn}可它的表情是\underline{我}从来没\underline{见}过的。\end{CJK}                                                    \\
                    % &                      & But it has a look \underline{I} have never \underline{seen} before.                       \\
                    % \cline{2-3}
                    & ref: \begin{CJK}{UTF8}{gbsn}可它的表情是\underline{我}从来没\underline{看}过的。\end{CJK}                                                    \\
                    % &                      & But it has a look \underline{I} have never \underline{looked at} before.                  \\
                    % \cline{2-3}
                    & \ \ \ \ \ \ \ But it has a look I have never seen before.\\
\hline
\multirow{9}{*}{\textbf{O}}  & src: \begin{CJK}{UTF8}{gbsn} 我班里有很漂亮的女同学，\underline{我}一见钟情。\end{CJK}                                               \\
                    % &                      & Air pollution has done some harm to \underline{citizens of Jin same}.         \\
                    % \cline{2-3}
                    & out: \begin{CJK}{UTF8}{gbsn}我班里有个很漂亮的女同学，\end{CJK}    \\
                    & \ \ \ \ \ \ \ \ \begin{CJK}{UTF8}{gbsn}\underline{她对我}一见钟情。\end{CJK}    \\
                    & \ \ \ \ \ \ \ \ There was a beautiful girl in my class. \\ & \ \ \ \ \ \ \ \ \underline{She} fell in love with \underline{me} at first sight.\\
                    % &                      & Air pollution has done some harm to \underline{citizens of Jin State}.        \\
                    % \cline{2-3}
                    & ref: \begin{CJK}{UTF8}{gbsn}我班里有位很漂亮的女同学，\end{CJK}                                               \\
                    & \ \ \ \ \ \ \ \ \begin{CJK}{UTF8}{gbsn}\underline{我对她}一见钟情。\end{CJK}                                               \\
                    % &                      & Air pollution has done some harm to \underline{ordinary citizens}.           \\
                    % \cline{2-3}
                    & \ \ \ \ \ \ \ \ There was a beautiful girl in my class. \\ & \ \ \ \ \ \ \ \ \underline{I} fell in love with \underline{her} at first sight.\\
\hline
\end{tabular}
\caption{\label{tab:examples}
Three examples for \textbf{G} and one for \textbf{O}. Label "src", "out" and "ref" means the source sentence, the output of one of our PLM-based ensemble strategies and the reference, respectively.}
\vspace{-1.0em}
\end{table}
% \begin{table}[h]
% \small
% \centering
% \begin{tabular}{|c|l|}
% \hline
% Type & Example\\
% \hline
% \multirow{9}*{\textbf{B}} & \begin{CJK}{UTF8}{gbsn}src: 我\underline{的}家附近有很多\underline{考式}补习班。\end{CJK}\\
% & \begin{CJK}{UTF8}{gbsn}hyp: 我家附近有很多\underline{考试}补习班。\end{CJK}\\
% & \begin{CJK}{UTF8}{gbsn}ref: 我\underline{的}家附近有很多\underline{考试}补习班。\end{CJK}\\
% \cline{2-2}
% & \begin{CJK}{UTF8}{gbsn}src: 我\underline{低幼儿童的时候}很想养狗。\end{CJK}\\
% & \begin{CJK}{UTF8}{gbsn}hyp: 我\underline{小时候}很想养狗。\end{CJK}\\
% & \begin{CJK}{UTF8}{gbsn}ref: 我\underline{小的时候}很想养狗。\end{CJK}\\
% \cline{2-2}
% & \begin{CJK}{UTF8}{gbsn}src: 可它的表情是从来没\underline{看}过的。\end{CJK}\\
% & \begin{CJK}{UTF8}{gbsn}hyp: 可它的表情是\underline{我}从来没\underline{见}过的。\end{CJK}\\
% & \begin{CJK}{UTF8}{gbsn}ref: 可它的表情是\underline{我}从来没\underline{看}过的。\end{CJK}\\
% \hline
% \multirow{3}{*}{\textbf{C}} & \begin{CJK}{UTF8}{gbsn}src: 空气污染已经对\underline{晋同}公民有一定的危害了。\end{CJK}\\
% & \begin{CJK}{UTF8}{gbsn}hyp: 空气污染已经对\underline{晋国}公民有一定的危害了。\end{CJK}\\
% & \begin{CJK}{UTF8}{gbsn}ref: 空气污染已经对\underline{普通}公民有一定的危害了。\end{CJK}\\
% \hline
% \end{tabular}
% \caption{\label{tab:examples}
% Three examples for \textbf{B} and one for \textbf{C}. Label "src", "hyp", and "ref" means the source sentence, the hypothesis sentence of our system, and the reference sentence.}
% \end{table}

\subsection{Discussion}
\paragraph{The insufficiency of GEC references.} In the outputs of PLM-based ensemble strategies, about 1/4 ("\textbf{G}") are automatically judged to be wrong according to the golden references, but indeed correct after human check. Actually, if we assume class \textbf{G} is also correct, the number of sentences corrected by PLM-based ensemble strategies (except edit-level ensemble) exceeds that by seq2seq-1, the best single model. 

This indicates that GEC references are not sufficient enough, even though datasets like NLPCC provide multi-references. Since artificially generating a correct sentence is much harder than judging a machine-generated sequence correct or not, continuously adding human checked results of PLM-ensemble 
%advanced GEC 
systems to the references may be a good solution to improve the quality and diversity of the GEC test data.  

\paragraph{The goal of GEC.}
This is a significant issue. Is it enough to just get a sentence rid of errors? Taking coding into example, can we say a piece of code "good" when all the "errors" are clear but pages of "warnings" are flashing? In "\textbf{Good}" samples, we compare the human references and automatically generated sentences, 
%We read some of the references of NLPCC-test and MuCGEC-dev 
and find many of references are only \textbf{correct} but not so \textbf{idiomatic}. On the other hand, many output sentences of PLM-based ensemble strategies are more natural and like native speakers. If a GEC system is aimed at helping overseas students with their language learning, for example, then idiomaticity should be taken into consideration. %In addition, if we assume class \textbf{B} is also correct, then the number of sentences corrected by PLM-based ensemble strategies (except edit-level ensemble) exceeds that by seq2seq-1, the best single model.

\paragraph{The over-correction of PLM-based models.}
About 1/10 of sentences generated in PLM-based ensemble ("\textbf{O}") are over-corrected, i.e., the model corrects a correct token and thus produces a wrong sentence. PLMs always choose the most fluent sentence with the lowest PPL, sometimes ignoring the original meaning of the source sentence. The over-correction of PLM-based generative models should be addressed in future work.   

%\paragraph{The essence of PPL.}
%PPL usually reflects the fluency of a sentence. When it comes to GEC, it is worth considering whether a sentence with a lower PPL is more "correct". As is shown in Table \ref{tab:statistical-results}, about 1/10 sentences are over-corrected. That is to say, PLMs always choose the most fluent sentence, whether it is properly corrected or not. Simply using PLMs to rate the correctness of several sentences may be inappropriate.

\section{Conclusion}
\label{section:conclusion}
%In this work, 
This paper introduces novel ensemble strategies for the GEC task by leveraging the power of pre-trained language models (PLMs). 
%Specifically, the proposed ensemble strategies exploit the perplexity score of PLMs to select the best prediction out of four different models.
We compare different strategies of model ensemble in CGEC. Surprisingly, PLM-based ensemble strategies do not benefit the system. This suggests that PPL and $F_{0.5}$ have diverging goals. According to our analysis, the insufficiency of references in GEC remains a major problem, which should be continuously improved in future work.

\section*{Acknowledgement}
This work is supported by the National Hi-Tech RD Program of China (No.2020AAA0106600), the National Natural Science Foundation of China (62076008) and the Key Project of Natural Science Foundation of China (61936012).

\section*{Limitations}
\label{sec:Limitations}
First, we don't use any single models without PLMs in their structures to carry out comparative experiments, even though few advanced models nowadays can get rid of PLMs. Second, because of the wrapping of fairseq, we don't have access to all the output probabilities of the single models and thus cannot apply the strategy of using the weighted sum of single models and PLMs used in \citet{junczys-dowmunt-etal-2018-approaching}. Third, while BERT-based PLMs are good at mask prediction, we haven't found a strategy to make use of that capacity without being embarrassed by conditional probability. Fourth, we carry out our experiments only on Chinese.

\section*{Ethics Statement}

\paragraph{About Scientific Artifacts.}
Since we focus on CGEC, all the code and tools are for the Chinese language and all data is in Chinese. All the scientific artifacts are used for GEC only. The artifacts provided by \citet{zhang-etal-2022-mucgec} are publicly available based on the Apache-2.0 license, on which we base our own codes and models.

\paragraph{About Computational Budget.}
We run all the experiments of model ensemble on an Intel$^\circledR$ Xeon$^\circledR$ Gold 5218 CPU. Processing times are shown in table \ref{tab:time}.

\begin{table}[h]
    \small
   \centering
   \begin{tabular}{lcc}
   \toprule
       \textbf{Strategy} & \textbf{MuCGEC-test} & \textbf{NLPCC-test} \\
   \toprule
       Traditional Voting & 1\textasciitilde2s & <1s \\
       Sentence-level & 25min & 6min \\
       Edit-level & 56min & 12min \\
       Edit-combination & 2.5h & 25min \\
   \bottomrule
   \end{tabular}
   \caption{Processing times of different ensemble strategies.}
    \label{tab:time}
\end{table}

\paragraph{About Reproducibility.}
All the experiments of model ensemble is completely reproducible when the PLMs are frozen (i.e., no matter how many times we run the experiments, the results are just the same).

\paragraph{About Human Annotators.}
Each of the annotators is paid \$20 per hour, above the legal minimum wage. The instructions are shown in Appendix \ref{sec:instructions}.

%Scientific work published at ACL 2023 must comply with the ACL Ethics Policy.\footnote{\url{https://www.aclweb.org/portal/content/acl-code-ethics}} We encourage all authors to include an explicit ethics statement on the broader impact of the work, or other ethical considerations after the conclusion but before the references. The ethics statement will not count toward the page limit (8 pages for long, 4 pages for short papers).

\bibliographystyle{acl_natbib}
\bibliography{anthology,custom}

\appendix
\section{Instructions for Human Annotation}
\label{sec:instructions}
The instructions for human annotators mentioned in Section \ref{sec:analysis} are as follows:

\begin{enumerate*}[itemjoin=\\\hspace*{\parindent}]
  \item You can see the data in "sample\_200.txt", which contains results of 200 sentences.
  \item Each sample contains several lines, including "Input" (the source sentence), "seq2seq-1", "Sentence-level", "Edit-level", "Edit-combination", and one or two "Reference" lines.
  \item You need to annotate the "seq2seq-1", "Sentence-level", "Edit-level" and "Edit-combination" lines according to the input and reference(s).
  \item To be specific, you should choose from the following four types. Exact (E): the output is fluent and correct, in line with the reference. Good (G): the output is fluent and correct but different with the reference, which indicates that the references are not sufficient enough. Over-corrected (O): the output is fluent but doesn't meet the original meaning of the source sentence. Wrong (W): the output has other problems that we don't care in this work.
  \item Thank you for your contributions!
\end{enumerate*}

\end{document}